\pdfoutput=1

\documentclass[11pt]{article}

\usepackage[final]{acl}

\usepackage[T1]{fontenc}
\usepackage[utf8]{inputenc}
\usepackage{times}
\usepackage{inconsolata}

\usepackage{amsmath}
\usepackage{amssymb}
\usepackage{latexsym}
\usepackage{eurosym}

\usepackage{booktabs}
\usepackage{multirow}

\usepackage{graphicx}
\usepackage{float}
\graphicspath{{./}{latex/}{figures/}}

\usepackage{microtype}

\usepackage{xcolor}

\definecolor{intruderred}{RGB}{200, 60, 50}

\title{FLAME: A New Dataset on FLemish Accounts of Momentary Experiences}

\author{
  Ratna Kandala\textsuperscript{1}, 
  Niels Vanhasbroeck\textsuperscript{2}, 
  Katie Hoemann\textsuperscript{1}\thanks{Corresponding author: \texttt{hoemann@ku.edu}} \\[4pt]
  \textsuperscript{1}University of Kansas, USA \quad
  \textsuperscript{2}University of Amsterdam, Netherlands \\[4pt]
  \texttt{\small ratnanirupama@gmail.com} \quad
  \texttt{\small n.d.p.vanhasbroeck@uva.nl} \quad
  \texttt{\small hoemann@ku.edu}
}

\begin{document}
\maketitle
\begin{abstract}
We introduce FLAME (FLemish Accounts of Momentary Experiences), a corpus of nearly 25,000 personal narratives in Belgian-Dutch (Flemish), collected through experience sampling to support Natural Language Processing (NLP) research on an underrepresented variety. Such everyday narratives are rich in culturally grounded themes, but their informal register and low-resource setting make thematic extraction hard. Comparing K-Means, LDA, and BERTopic, we find that human evaluation favors BERTopic, which produces the most coherent, culturally resonant topics. FLAME, thereby, offers a new resource for studying everyday language use in a low-resource variety.
\end{abstract}

\section{Introduction}
Language is used continuously in the course of everyday life by people when they share information on what they are doing, where they are located, and how they feel about the events unfolding around them. Open-ended personal narratives of this kind are a rich source of information to study human behavior in relation to their activities, social interactions, feelings, and their surroundings or contexts. For NLP, such daily narratives offer something curated corpora rarely do-language sampled close to the experience it describes, from the same individuals, repeatedly over time. Yet despite the rapid growth in large-scale datasets and benchmarks, resources that capture open-ended first-person descriptions of everyday experiences remain scarce, and even more for regional and underrepresented language varieties \cite{joshi2020state}. 

The availability of dedicated datasets has been central to progress in NLP research. Many resources have driven advances across tasks such as comprehension \citep{rajpurkar2016squad}, natural language inference \citep{williams2018broad}, and general language understanding \citep{wang2018glue}, among others. However, the language material has mainly been drawn from a narrow set of domains, including Wikipedia, online discussions, news, or other pre-existing texts. The growing number of NLP datasets therefore does not translate into comparable coverage of the language people actually use to describe their own lives as they live them.

According to \cite{sultana2022narrative}, narrative resources in NLP fall into a small number of recurring types. For instance, many are artificial or crowdsourced stories - for example, the five-sentence commonsense narratives of ROCStories \cite{mostafazadeh2016} which support specific technical tasks but carry little social grounding. Others are curated corpora of novels, news, biographies, or social-media posts; these are authored in contexts separate from the research and gathered retrospectively, and are thus analyzed at a remove from the situations that produced them. Massive pretraining corpora \cite{gao2020pile} extend this further, mixing documents from unrelated domains at a scale that precludes documentation of their contents \cite{gebru2021datasheets,bender2021parrots}. Naturalistic narratives elicited for the study itself are comparatively rare \cite{sap2020}, and rarer still are resources that sample the \emph{same} speakers repeatedly, in situ, as their experiences unfold.

This gap matters specifically for personal narratives. Unlike text written for encyclopedic, journalistic, or social-media audiences, accounts of everyday experience are anchored in an individual's immediate situation. They interleave mundane activity with personally salient events, shifts in affect, references to specific people and places, and culturally or regionally particular expressions. Crucially, the same person produces markedly different accounts from one moment to the next, so that repeated sampling exposes within-person linguistic variation that cannot be reconstructed from documents collected once, retrospectively, or scraped from the web. For NLP, data with these properties enable questions that static corpora cannot easily support: modeling how an individual's language evolves over time, grounding free-text descriptions in self-reported experience, and studying naturalistic, dialectal, and code-mixed usage as it actually occurs.
A further limitation of existing narrative data is that the author's own perspective is typically absent. Affective and other internal states, when labeled at all, are usually inferred post hoc by third-party annotators rather than reported by the person who lived the experience, a practice that risks imposing observers' interpretations onto states known only to the author, and that carries documented hazards when internal states are treated as objectively classifiable \cite{stark2018}. Narratives collected close to the experience, and paired with the narrator's own account of how they felt, preserve a positionality that experience-distant labeling tends to erase.

Efforts to document language in genuinely everyday settings underscore the value of this perspective. The ORD corpus, for instance, was built to capture Russian everyday communication \cite{asinovsky2009ord,bogdanova2016ord}, showing that naturally occurring communication surfaces phenomena poorly represented in more formal corpora. Large-scale resources centered on repeated personal accounts of everyday experience nonetheless remain uncommon, and rarer still for varieties already underrepresented in NLP \citep{wu2025mmerc}.

Belgian-Dutch (Flemish) is a case in point. Substantial resources for Dutch and Flemish exist, but they were built for different communicative settings and purposes. The Spoken Dutch Corpus (CGN) assembled roughly ten million words of contemporary standard Dutch as spoken in the Netherlands and Flanders \cite{oostdijk2000spoken}, an invaluable resource for speech and language technology, but broader in aim than documenting individuals' repeated personal narratives. The Netlog Corpus collected Flemish-Dutch chat posts and user profiles from a Belgian social network \cite{kestemont2012netlog}, providing a gold standard for normalizing informal Flemish Internet language; its material, however, is produced for an online audience rather than elicited during ongoing experience. EmotioNL annotates Dutch tweets and reality-television captions for categorical emotions and for valence, arousal, and dominance \cite{debruyne2021emotionl}, advancing Dutch affect detection but likewise drawing on specific media and online contexts rather than longitudinal sampling of daily life.

We address this gap by introducing FLAME (FLemish Accounts of Momentary Experiences), a longitudinal corpus of nearly 25,000 daily personal narratives produced by 102 native Belgian-Dutch speakers over a 70-day experience-sampling study. Participants responded four times per day to an open-ended prompt asking what was happening and how they felt about it, using either a one-minute voice message or a short typed response. The resulting corpus captures informal, contextually grounded language describing everyday experiences as they unfold. To our knowledge, FLAME is the first large-scale Belgian-Dutch corpus of personal narratives collected prospectively and in situ through experience sampling; its novelty lies in the naturalistic, longitudinal data collection procedure, and within-person manner in which its language was elicited.

We then provide an initial characterization of the corpus's thematic structure. Because these are open-ended accounts collected without a predetermined label scheme, we adopt an exploratory, discovery-oriented approach using three widely used methods:  Term Frequency-Inverse Document Frequency (TF-IDF) \cite{salton1988term, ramos2003using}  based KMeans clustering approach \cite{macqueen1967some, lloyd1982least, berkhin2006survey, sinaga2020unsupervised}, the probabilistic LDA model \cite{blei2003latent, shin2025comparison}, and BERTopic \cite{grootendorst2022bertopic}, assessing which best captures the coherent and culturally grounded themes that FLAME contains. Critically, this validation moves beyond automated coherence metrics alone, which we argue can be insufficient or even misleading for embedding-based models on narrative data. We therefore complement automated metrics with a human evaluation of topic interpretability and cultural relevance. Our results show that BERTopic outperforms LDA on three of four coherence metrics, LDA leads only on $C_v$, and produces the most coherent and culturally resonant topics, demonstrating that contextual embeddings are critical for this type of data, and that human-centered evaluation is indispensable when working with low-resource, narrative-rich corpora.

The contributions of this work are threefold:
(a) We introduce FLAME, a longitudinal corpus of nearly 25{,}000 naturalistic personal narratives collected by 102 Dutch-speaking adults residing in Belgium.
(b) We characterize this corpus to select an appropriate method for open-ended thematic analysis of in-situ narratives.
(c) We provide an exploratory thematic characterization of everyday Flemish narrative, illustrating a range of activities, experiences, and culturally grounded themes represented in FLAME and its potential for further NLP research on naturalistic language.

The remainder of this paper is structured as follows: Section 2 discusses the data collection and data preparation methodology employed for this study, including a few illustrative examples. Section 3 provides an overview of the methodology to validate this dataset. Section 4 presents the results. Section 5 discusses the findings, Section 6 presents the limitations, and Section 7 concludes the study.

\vspace{-1 mm}

\section{Data and Data Collection}
The dataset consists of open-ended responses of daily narratives in Flemish with a total of 24,854 texts. Participants $(N=102, age: 18-65, M=26.47, SD=8.87)$ were recruited via flyers, online posts, and word-of-mouth. Eligible individuals were native Dutch speakers living in Belgium, had to be at least 18 years old, and had to be in possession of a working smartphone. Over 70 days, participants received four prompts per day via a smartphone app called m-Path \cite{mestdagh2023mpath}, responding to the question: \textit{"Wat speelt er nu of sinds de vorige beep, en hoe voel je je daarover?”
} (“What is going on now or since the last prompt, and how do you feel about it?”). They provided either a one-minute voice message or a 3-4 sentence text response. 

\subsection{Data Preparation}

\noindent Recorded descriptions were automatically transcribed using a Dutch automatic speech recognition (ASR) system developed in-house \cite{tamm2024weakly}. This model was trained on approximately 270 hours of Flemish Dutch speech from the Spoken Dutch Corpus  \cite{oostdijk2000spoken}, following protocols similar to the baseline described by \cite{PonceletVanHamme2023JointASR}. The resulting transcripts contain system-generated punctuation and adhere to the orthographic transcription conventions of the Spoken Dutch Corpus, which include detailed annotation tags for foreign words, dialect, disfluencies, and other phenomena. Subsequent preprocessing removed these annotation tags and non-content markers, retaining only the core textual content for further analysis. For additional details on participant recruitment and study procedures, see Appendix.

\subsection{Illustrative Examples}

To illustrate the nature of the transcripts, Table \ref{tab:examples} presents a few representative examples from the dataset. Each entry includes the raw Dutch text, its English translation, the valence score, and the word count. These examples reflect the naturalistic, diary-style language characteristic of the corpus, ranging from workplace experiences to social and leisure activities.

\begin{table*}[!ht]
\centering
\small
\begin{tabular}{p{5cm} p{5cm} c c}
\hline
\textbf{Raw Text (Dutch)} & \textbf{Translation (English)}  & \textbf{Words} \\
\hline
net een vermoeiende maar wel echt goede groepsessies gehad op mijn studentenjob. ik voel me wat overprikkeld, maar op zich voel ik me wel echt goed. nu ben ik bij vrienden voor de verjaardag van een vriendin dus ben wel blij om ze terug te zien.
 & Just had exhausting but really good group sessions at my student job. I feel a bit overstimulated, but overall I feel really good. Now I'm with friends for a girlfriend's birthday, so I'm happy to see them again.
 & 46 \\
\hline
Ik heb net een hele tijd in de auto gezeten en dat heeft me wat suf gemaakt. Het doet me uitkijken naar de wandeling die we gaan maken. Autoritten maken me snel overprikkeld dus ik hoop dat dat snel wegtrekt door buiten te zijn.
 & I've just been in the car for long time and that made me a bit groggy. Makes me look forward to the walk we're going to take. Car rides tend to overstimulate me, so I hope being outside will help me snap out of it soon.
& 44 \\
\hline
ik ben in een meeting momenteel op het werk. het is best saai. ik ben ook mijn simulaties aan het checken, en ze zien er een beetje raar uit. ik ben een klein beetje gedemotiveerd, maar ik ga proberen om mijn motivatie terug te vinden. hopelijk lukt dat
 & I'm in a meeting at work right now. It's pretty boring. I'm also checking my simulations, and they look a bit strange. I'm a little demotivated, but I'm going to try to get my motivation back. Hopefully it'll work out.
& 48 \\
\hline
ik heb nog in de avond in het atelier geweest. gewoon wat tijd met mezelf en heb verschillende kommen gemaakt. dat deed goed, het kriebelde al de hele dag om te creëren. ik heb ook met, mijn lief gebeld en afgesproken voor morgen. het deed goed om met hem te bellen. nu ben ik met m'n ouders tv aan het zien
 & I spent some time in the studio this evening. Just some time to myself, and I made a few bowls. That felt good - I'd been itching to create all day. I also call my boyfriend and made plans for tomorrow. It was nice to talk to him. Now I'm watching TV with my parents.
& 62 \\
\hline
\end{tabular}
\caption{Representative examples from the FLAME dataset}
\label{tab:examples}
\end{table*}

\section{Methodology}

\subsection{Data Analysis}

\subsubsection{Preprocessing}
The raw texts underwent a multi-step preprocessing pipeline: (i) removal of dataset-specific annotation tags and author-identifying references (ii) lemmatization using the Stanza \cite{qi2020stanza} model for Dutch, and (iii) filtering of documents shorter than 15 words. A custom, corpus-specific stopword list was also applied \footnote{https://github.com/ratnakandala/FLAME}. The final processed corpus contains 24,752 documents.

%

To quantitatively define 'medium-length,' we report word count statistics for both unfiltered and filtered corpora. Prior to filtering, transcripts had a mean of 109.34 words (SD = 70.29), with a median of 101 words (range: 1–965). After applying the 15-word minimum filter, the mean was 109.84 words (SD = 70.11), with a median of 102 words (range: 10–965). The minimal difference between unfiltered and filtered statistics indicates that very short responses were rare, and that medium-length responses are a consistent feature across the study population.

\subsubsection{Model Selection, Configuration, and Hyperparameter Optimization}

We conduct a comparative analysis of KMeans, LDA, and BERTopic on FLAME. \footnote{Regarding NMF, since it can be applied to the same TF-IDF feature space used by KMeans, we scoped our comparison to a single representative TF-IDF-based baseline (KMeans), alongside probabilistic LDA and embedding-based BERTopic.} Our evaluation framework is twofold, comprising a suite of automated topic coherence and diversity metrics, supplemented by human evaluation to assess topic interpretability. 

\subsubsection*{KMeans Clustering}
For the KMeans baseline, documents were vectorized using TF-IDF, a classical, non-semantic feature space following the approach of \cite{kamiloglu2025what}. To establish a robust baseline, we conducted a systematic exploration of vectorizer parameters, testing various combinations of min\_df and max\_df, including both absolute counts and proportional thresholds. The final configuration reported in this study (min\_df=0.01, max\_df=0.9) was selected to generate a compact vocabulary focused on high-frequency terms, resulting in a document-term matrix (DTM) of size 24,725 $\times$ 326. Sublinear term frequency scaling \cite{manning2008introduction} was also applied to reduce the influence of high-frequency terms. For the clustering step, the optimal number of clusters (k) was determined via Silhouette analysis \cite{rousseeuw1987silhouettes} over a range of $k=2$ to $200$, which indicated a peak score at $k=128$ (see Appendix). KMeans clustering was performed using Euclidean distance as the distance metric, which is the standard metric used in KMeans optimization \cite{lloyd1982least}. The entire pipeline was implemented using Python’s scikit-learn \cite{scikit-learn} library with fixed random seeds to ensure reproducibility.

\subsubsection*{LDA}
\noindent For the LDA baseline, we used the scikit-learn implementation \cite{scikit-learn}. We tuned the number of topics($k$) (ranging from 5 to 150), $\alpha$ (\texttt{doc\_topic\_prior}), and $\beta$ (\texttt{topic\_word\_prior}) over [0.01, 0.1, 1.0, 0.5]. Model selection was guided by coherence scores computed with the Gensim coherence model \cite{rehurek2011gensim}.


\subsubsection*{BERTopic}
We systematically tuned BERTopic’s hyperparameters using \emph{jina-embeddings-v3} to optimize topic coherence and diversity. 
For dimensionality reduction, we tested \texttt{umap\_n\_neighbors} with values [10, 15, 20, 25, 30] and \texttt{umap\_min\_dist} with values [0.0, 0.1]. 
For clustering, we evaluated \texttt{hdbscan\_min\_cluster\_size} with values [5, 10, 15]. 
For the model, we tested \texttt{min\_topic\_size} with values [5, 10, 15] and \texttt{nr\_topics} with the value [\texttt{auto}]. 
Initial clustering with BERTopic labeled 59\% of documents (n = 14{,}600) as outliers. 
To mitigate this, we employed the \emph{Approximate Distribution Method}, which probabilistically assigns outliers to the nearest topic based on cosine similarity in embedding space \cite{grootendorst2022bertopic}. 
This reduced outliers to 7.56\% (n = 1{,}872) for the best BERTopic $C_v$ score observed.

\vspace{1 mm}
\noindent \textbf{Embedding Model Selection for BERTopic}
To identify the optimal sentence-level embeddings for Flemish, we evaluated three multilingual models: (a) \emph{robert-2022-dutch-sentence-transformers} \cite{netherlands_forensic_institute_2024}: A Dutch-specific model fine-tuned for semantic similarity task (b) \emph{gte-multilingual-base} \cite{zhang2024mgte}: A general-purpose multilingual encoder optimized for long-context retrieval (c) \emph{jina-embeddings-v3} \cite{sturua2024jinaembeddingsv3multilingualembeddingstask}
: A multilingual model leveraging Task LoRA for adaptable semantic representation. Qualitative assessment of the resulting topics indicated that \textit{jina-embeddings-v3} produced the most coherent themes for our specific corpus. We attribute this to its XLM-RoBERTa architecture augmented with Rotary Position Embeddings and Task LoRA adapters. Consequently, all BERTopic experiments utilize this model, which generates 1,024-dimensional embeddings via the SentenceTransformers library \cite{reimers2019sentence}. We then utilized Uniform Manifold Approximation and Projection (UMAP) \cite{mcinnes2018umap} to reduce these embeddings to a 5-dimensional space.

\subsection{Evaluation Metrics}

To provide a comprehensive comparison, we evaluate the models on both automated metrics and human judgments.

\subsubsection{Automated Metrics}
We employ a suite of four standard topic coherence metrics \cite{mimno2011optimizing} to evaluate the interpretability and semantic consistency of topics generated by BERTopic and LDA. $C_v$ measures coherence based on a combination of Normalized Pointwise Mutual Information (NPMI) and cosine similarity over a sliding window (size 110), closely aligning with human judgments of word co-occurrence \cite{roder2015exploring}. $C_\text{npmi}$ is a normalized PMI score robust to frequency biases, scaled between -1 (dissociation) and +1 (perfect association), making it suitable for comparing embedding-based models (where semantic similarity extends beyond raw co-occurrence) and count-based models. As a normalized and stable variant of PMI, it is less susceptible to biases from word frequency, making it a valuable metric for comparing embedding-driven models like BERTopic (which prioritize semantic similarity over raw co-occurrence) against count-based models like LDA. $U_\text{mass}$ evaluates topic quality based on the log conditional probability of word co-occurrences, favoring probabilistic models like LDA. Finally, $U_\text{uci}$ uses raw PMI as a baseline co-occurrence measure. We also compute topic diversity, defined as the percentage of unique words in the top-10 words of all topics.\\

\vspace{-3.5 mm}

\subsubsection{Human Evaluation of the Topics} 
To complement the automated coherence metrics, we conducted a human evaluation study to assess the semantic coherence of the topics generated by each model. Two human annotators, both Dutch speakers, participated in a word intrusion task. This task is used to probe whether topic words form a semantically cohesive group by asking annotators to identify a word that does not semantically belong \cite{chang-etal-2009-reading, bhatia-etal-2018-topic}, and is a widely used method for evaluating topic coherence.

For each of the three models (KMeans, LDA, BERTopic), we randomly sampled 40 topics. For each topic, we constructed a six-word set comprising the top five words associated with the topic and a sixth "intruder" word. Intruder words were drawn from high-probability terms of unrelated topics within the same model.  Specifically, an intruder was sampled with uniform probability from the top 10 words of a randomly selected, unrelated topic. This method ensures that the intruder is a plausible, high-frequency word in its own right, making the task a robust test of the target topic's semantic coherence. The order of the sixth (intruder) word was randomized for each trial to minimize position bias. Annotators were instructed to independently identify the single word in each set that did not semantically fit with the others. We report two evaluation metrics in this regard: (i) Topic Coherence Accuracy: This constitutes the proportion of topic sets in which the annotator correctly identified the intruder word. This serves as an estimate of the interpretability and semantic coherence of the topics \cite{chang-etal-2009-reading} (ii) Inter Annotator Agreement (IAA): To assess annotation reliability, we computed Krippendorff's Alpha $\alpha$ \cite{krippendorff2004content, artstein_inter-coder_2008}, a standard metric for agreement on categorical judgments. Here, agreement reflects whether annotators identified the same intruder word in each set. High agreement indicates consistent judgments about topic coherence across annotators. Figures illustrating annotator accuracy and inter-annotator agreement are provided in the appendix.

\section{Results and Analysis}
Our analysis compares the embedding-based BERTopic model against a probabilistic LDA baseline and a TF-IDF-based KMeans baseline on a corpus of over 24,000 Flemish daily narratives. The results reveal a significant divergence between automated coherence metrics and human-perceived topic quality, highlighting the limitations of traditional evaluation paradigms for semantically-aware models on structurally diverse data.

\subsection{Quantitative Analysis}
The quantitative evaluation, summarized in Tables \ref{tab:cvscores} and \ref{tab:coherences}, presents a nuanced picture. In line with prior work on topic models applied to short or medium-length texts, performance of both BERTopic and LDA peaked at an optimal number of topics before declining, particularly in datasets with limited word counts per document \citep{aggarwal2012survey, muthusami2024investigating, mutsaddi2025bertopic}.

When comparing the best-performing configurations of each model, we found that LDA produced higher  $C_v$ topic diversity scores than BERTopic (Table \ref{tab:cvscores}). Specifically, LDA achieved 
a $C_v$ score of 0.5430 for 120 topics, and BERTopic achieved 0.3412 for 76 topics. We focused on $C_v$ as our principal coherence measure because it is widely used in topic modeling studies. The higher score for LDA suggests stronger word-level topical coherence, 
likely due to its reliance on explicit term co-occurrence statistics, 
which aligns with traditional interpretability expectations \citep{roder2015exploring}. 
On the other hand, the lower $C_v$ score of BERTopic 
could reflect its dependency on contextual embeddings, 
which prioritize semantic similarity over lexical overlap, 
a known limitation of PMI-based metrics in embedding-driven models \citep{grootendorst2022bertopic}.
Table~\ref{tab:cvscores} presents the $C_v$ scores for selected topic counts with both models. 
Topic counts with very poor coherence scores have been omitted.

\begin{table}[ht]
\centering
\begin{tabular}{llr}
\hline
\textbf{Model} & \textbf{No.\ of Topics} & \textbf{$C_v$ Score} \\
\hline
\multirow{4}{*}{\small BERTopic}  & 60  & 0.296 \\
                                  & 76  & \textbf{0.341} \\
                                  & 110 & 0.337 \\
                                  & 117 & 0.337 \\
\hline
\multirow{4}{*}{LDA}              & 76  & 0.519 \\
                                  & 100 & 0.505 \\
                                  & 120 & \textbf{0.543} \\
                                  & 148 & 0.542 \\
\hline
\end{tabular}
\caption{$C_v$ scores for selected topic counts with BERTopic and LDA.}
\label{tab:cvscores}
\end{table}

However, BERTopic produced higher scores on the other coherence metrics 
($C_\text{npmi}$, $U_\text{mass}$,  $U_\text{uci}$) than LDA, suggesting that $C_\text{npmi}$’s stability 
accommodated BERTopic’s embedding-based semantics better than raw PMI \citep{lau2014machine}. 
Furthermore, $U_\text{mass}$, which penalizes rare word pairs, 
may align better with BERTopic’s ability to cluster medium-length texts 
without overfitting to sparse co-occurrences. 
Table~\ref{tab:coherences} compares the scores of the best BERTopic and best LDA models 
across four coherence metrics and topic diversity.

\begin{table*}[ht]
\centering
\begin{tabular}{lcccccc}
\hline
\textbf{Model} & \textbf{No.\ of Topics} & \boldmath\(C_v\) & \textbf{$C_\text{npmi}$ } & \textbf{$U_\text{mass}$} & \textbf{$U_\text{uci}$} & \textbf{Topic Diversity} \\
\hline
BERTopic & 76  & 0.3412  & \textbf{-0.1619} & \textbf{-12.2167} & \textbf{-5.6599} & 0.8455 \\
LDA      & 120 & \textbf{0.543}   & -0.21   & -16.39   & -5.93   & \textbf{0.9675} \\
\hline
\end{tabular}
\caption{Topic coherence and topic diversity for the best hyperparameter configurations.}
\label{tab:coherences}
\end{table*}

In terms of topic diversity, LDA achieved a higher score of 0.967 for 120 topics, 
compared to BERTopic’s 0.845. We interpret these findings to indicate 
that LDA’s Dirichlet prior encourages distinct topic distributions, 
while BERTopic’s HDBSCAN clustering allows overlapping semantic themes, 
which could reduce diversity but potentially capture nuanced relationships (see Appendix for details on Topic Distribution).

The KMeans baseline, optimized using geometric metrics like the Silhouette score, favored a high number of clusters (k=128) (see Appendix). However, this granularity resulted in a marked decline in topic coherence and interpretability, underscoring the potential mismatch between geometric partitioning objectives and the goal of extracting semantically meaningful themes \citep{aggarwal2012survey}.

\subsection{Qualitative Analysis: Best KMeans vs.\ Best LDA vs.\ Best BERTopic}
While LDA achieved the highest $C_v$ score, human evaluation revealed significant discrepancies in semantic coherence and contextual relevance, particularly evident in the morphologically rich and regionally specific Flemish corpus. BERTopic consistently generated thematically cohesive and culturally resonant topics, while LDA and KMeans struggled with semantic fragmentation and noise. 
The themes identified by BERTopic were specific and interpretable, for example, differentiating everyday routine ('workday routine', 'planning and communication', and 'studying') from activities ('horse riding', 'travel and outdoor recreation', and 'film evenings'), "Chiro" (Belgium's largest youth organization), "rains" (as it frequently rains in Belgium) and overall emotional and mental state ('headaches and migraine-related pain', 'academic stress and assignments').


BERTopic's superior ability to capture relevant themes can be illustrated with the topics related to "studying and academic life" (Table \ref{tab:studytopics}). It successfully captures
"\textit{aula\_vriend}" (\emph{lecture hall friend}) and "\textit{bibliotheek}" (\emph{library}), while the LDA model conflated academic terms with noisy co-occurrences like “\textit{bus}” and “\textit{toekomst}” (\emph{future}), likely due to its reliance on document-level word distributions,
which struggle with sparse co-occurrence patterns in medium-length texts.
KMeans clustered high-frequency but semantically unrelated verbs “\textit{eten}” (\emph{to eat}) and “\textit{slapen}” (\emph{to sleep}), failing to isolate domain-specific themes.

\begin{table}
  \centering
  \small
  \resizebox{\columnwidth}{!}{%
  \begin{tabular}{|l|p{6cm}|}
    \hline
    \textbf{Model} & \textbf{Top Words} \\
    \hline
    BERTopic & 
      \textbf{studeren} (to study), bibliotheek (library), tevreden (satisfied), 
      aula\_vriend (lecture hall friend), studie\_tijd (study time), 
      leeszaal (reading room), bib (library), dezeochtend (this morning), basically \\ \hline

    LDA & 
      \textbf{studeren} (to study), bus, snappen (to understand), 
      interessant (interesting), deadline, toekomst (future), 
      af\_sluiten (to wrap up), practicum (lab session), focus \\ \hline

    KMeans & 
      \textbf{studeren} (to study), \underline{goed} (good), \underline{vandaag} (today), 
      \underline{eten} (to eat), \underline{beginnen} (to begin), \underline{rest}, 
      \underline{dag} (day), ver, \underline{moe} (tired), tijd (time), oké (okay),
      \underline{weten} (to know), slapen (to sleep), morgen (tomorrow), 
      \underline{zin} (motivation) \\ \hline
  \end{tabular}
  }
  \caption{Topics identified by BERTopic, LDA, and KMeans related to ``studying and academic life.''}
  \label{tab:studytopics}
\end{table}

This pattern persisted in domains like “fitness” (Table~\ref{tab:table7}). 
BERTopic cohesively grouped terms such as “\textit{fitnessen}” (\emph{to work out}), “\textit{gym},” and “\textit{kracht\_training}” (\emph{power training}),
while LDA’s topic included semantically discordant words such as “\textit{broer}” (\emph{brother}) and “\textit{baby}.”
KMeans conflated fitness with unrelated daily activities such as “\textit{eten}” (\emph{to eat}) and “\textit{moe}” (\emph{tired}). Another trend observed in our KMeans analysis is that many common words (underlined in the tables) reappeared across clusters.

\begin{table}[ht]
  \centering
  \small
  \resizebox{\columnwidth}{!}{%
  \begin{tabular}{|l|p{6cm}|}
    \hline
    \textbf{Model} & \textbf{Top Words} \\
    \hline
    BERTopic & 
      \textbf{fitnessen} (to work out), fitness, workout, gym, oefening (exercise), 
      trainen (to train), joggen (to jog), kracht\_training (power training), 
      sport\_les (sports class), sport\_school (sports school) \\ \hline

    LDA & 
      \textbf{fitnessen} (to work out), time, baby, start, haasten (to hurry), 
      toe\_voegen (to add), quality, model, aanpassing (adjustment) \\ \hline

    KMeans & 
      \textbf{fitnessen} (to work out), \underline{goed} (good), \underline{eten} (to eat), 
      \underline{dag} (day), \underline{vandaag} (today), \underline{weten} (to know), 
      leuk (nice/fun), \underline{zin} (mood), vanavond (tonight), \underline{moe} (tired), 
      \underline{beginnen} (to begin), studeren (to study), \underline{rest}, 
      werken (to work), proberen (to try) \\ \hline
  \end{tabular}
  }
  \caption{Topics identified by BERTopic, LDA, and KMeans related to ``fitness.''}
  \label{tab:table7}
\end{table}

\subsection{Human Evaluation Results} 
Our evaluation showed that BERTopic achieved notably higher precision (Annotator 1: 95.0 \%, Annotator 2: 87.5 \%) compared to KMeans (Annotator 1: 60.0 \%, Annotator 2:  52.5.0 \%) and LDA (Annotator 1: 20.0\%, Annotator 2: 25.0\%). This substantial gap indicated that topics produced by BERTopic have better interpretability. 

To assess the consistency and ensure the reliability of annotators' judgments, we also computed the Inter-Annotator Agreement (IAA) using Krippendorff's Alpha \cite{krippendorff2004content}. Agreement was high for BERTopic ($\alpha = 0.874$), further reinforcing confidence in the model's topic coherence. In contrast, moderate agreement levels were observed for KMeans ($\alpha = 0.545$) and LDA ($\alpha = 0.547$), suggesting relatively lower coherence and greater ambiguity in topics generated by these models. These results highlight BERTopic's strength in producing semantically coherent and interpretable topics.

\section{Discussion}
\label{sec:discussion}

This study's findings reveal a critical tension in topic model evaluation: the divergence between automated coherence metrics and human judgments of topic quality, particularly when applied to a structurally diverse corpus of Flemish daily narratives. 

The superior $C_v$ score achieved by LDA, juxtaposed with its poorer performance in human evaluation, points to a systemic limitation of metrics that rely on surface-level co-occurrence. As evidenced by the qualitative analysis, LDA’s inclusion of frequent but semantically irrelevant co-occurrences 
(e.g., “\textit{bus}” near “\textit{studeren}” [\emph{to study}]) 
could reflect its bias toward “syntagmatic associations” (statistical proximity) 
rather than true “paradigmatic relevance” (thematic consistency). 
In contrast, BERTopic’s embeddings prioritize contextual relationships 
(e.g., “\textit{bibliotheek}” [\emph{library}] $\leftrightarrow$ “\textit{studie\_tijd}” [\emph{study time}]), 
aligning with human intuition \citep{lau2014machine}. This becomes especially problematic when one uses these methods to extract topics that are specific to a particular culture. Within this dataset, for example, BERTopic was able to identify Flemish-related topics related to the youth organization Chiro, national holidays like Christmas, and the weekly commute of university students from Leuven to home and back. Our results thus underscore the necessity of adopting hybrid evaluation frameworks that integrate human-in-the-loop validation, especially for under-resourced languages and non-standard text genres \citep{muthusami2024investigating}.

This discrepancy strongly suggests that an over-reliance on automated metrics like $C_v$ can be misleading for evaluating embedding-based models, whose strengths lie in capturing semantic nuances that bag-of-words approaches miss. Furthermore, the high inter-annotator agreement for BERTopic (Krippendorff's $\alpha$ = 0.874), compared to the moderate agreement for the baselines, indicates that BERTopic's topics are not only more coherent but also more clearly and consistently interpretable. 

Beyond the choice of topic modeling approach, our results also point to the importance of model configuration. The best-performing BERTopic configuration combined a multilingual sentence transformer for contextual embeddings with HDBSCAN for density-based clustering and c-TF-IDF for topic representation, underscoring that no single component alone drives performance. Crucially, we attribute BERTopic's success not merely to its use of contextual embeddings, but to the synergy between those embeddings and HDBSCAN's density-based clustering strategy. Unlike partition-based methods such as KMeans, which force every document into a cluster, HDBSCAN is specifically designed to handle noise and outliers, a property that proves essential for FLAME, where 59\% of documents were classified as outliers. This high outlier rate reflects the inherently open-ended and thematically diffuse nature of daily personal narratives, and suggests that density-based clustering is better suited to this kind of data than partition-based alternatives.

Finally, our results with the K-Means baseline warrant discussion. While prior work demonstrated the efficacy of KMeans on Dutch texts \citep{kamiloglu2025what}, our experiments did not replicate this success, instead yielding generic and incoherent clusters.  We posit that this is not a failure of the algorithm itself, but rather a limitation of the TF-IDF feature space when applied to the unstructured, highly variable style of personal narratives. The vocabulary pruning inherent in TF-IDF, even with systematic tuning, appears insufficient to create a vector space where thematically distinct narratives form separable geometric clusters, resulting in the observed low-quality output. 


\vspace{- 1.5 mm}

\section{Limitations}
Our study, while providing a focused analysis of topic modeling on a unique dataset, is subject to several limitations that frame the scope of our conclusions. First, the generalizability of our findings is constrained by our use of a single, highly specific corpus: daily personal narratives from a predominantly young adult, Dutch-speaking population. The linguistic characteristics of this genre: informal, unstructured, and reflective differ significantly from other text types like news articles or formal documents. Consequently, the relative performance of BERTopic and the baseline models may not directly translate to other domains, languages, or demographic groups. The topics themselves, such as "studying" and "internships," reflect the life stage of our participants and are not representative of the entire population. Second, our methodology is dependent on the quality of upstream NLP components, which have known limitations for non-standard language variants. We selected the \textit{jina-embeddings-v3} model after a qualitative comparison, but even state-of-the-art multilingual models may not optimally capture the dialect-specific nuances of Flemish compared to a hypothetical native model. Furthermore, our preprocessing pipeline relied on Stanza for lemmatization. Errors or inconsistencies in lemmatizing dialectal or informal terms (e.g., \textit{dezeochtend}) can introduce noise that disproportionately affects the performance of count-based models like LDA and the TF-IDF-based KMeans, potentially impacting the fairness of our comparison. Finally, while this study advocates for the superiority of BERTopic's qualitative output, we acknowledge the inherent subjectivity of topic model evaluation. Our conclusions are based on a combination of automated metrics and human judgment, yet both have constraints. Automated coherence scores are imperfect proxies for human interpretation, and human evaluation, while essential, is difficult to scale and can be influenced by the specific task design and the evaluators' own biases.

\section{Conclusion and Future Work}
\label{sec:conclusion}
We introduce FLAME, a new corpus of nearly 25k daily personal narratives in Flemish, and explored which topic modeling approach is best suited to uncover the latent themes embedded in data of this kind. Our findings demonstrate that while traditional models like LDA can achieve higher scores on co-occurrence-based metrics such as $C_v$, BERTopic generates qualitatively superior topics that are more semantically coherent and culturally resonant. This work highlights a critical challenge in the field: the potential for automated metrics to be misleading when evaluating modern, embedding-based models on morphologically rich and context-dependent text, underscoring the need for hybrid evaluation frameworks and human-in-the-loop validation.

Beyond its methodological contributions, FLAME opens up new possibilities for computational social science research on underrepresented language varieties \cite{kandala2026syntactic}. Traditional text analysis in this domain often relies on lexicon-based tools like LIWC \cite{pennebaker2001linguistic}, which are limited by their static, non-contextual nature. FLAME, combined with contextual topic modeling approaches validated here, offers a more nuanced, data-driven lens into the thematic content of everyday personal experience in a culturally specific community, one that has been largely invisible in NLP research to date. 

The corpus also supports a range of tasks beyond topic modeling. The self-reported valence annotations already present in the dataset provide a natural starting point for affect-oriented work, such as valence prediction and modeling how emotion is linguistically constructed within everyday contexts  \cite{mohammad2022ethics, kandala2025llms}. Because responses were collected in two modalities, FLAME further enables comparison of the themes and language elicited from spoken versus typed responses. More broadly, as a naturalistic, in-situ resource in a low-resource variety, FLAME complements ongoing efforts to build evaluation resources for Flemish and Dutch, such as community benchmarking initiatives \cite{flembench} and established repositories for Dutch-language materials.\footnote{E.g., the Instituut voor de Nederlandse Taal \cite{int_taalmaterialen} and CLARIN \cite{clarin_nl}}. It can thereby serve future work on culturally grounded NLP and the computational analysis of everyday human experience.


\section*{Ethical Considerations}
  All participants provided informed consent and were fully briefed on the study’s procedures, including the secure collection of self-reported responses and sensor data. Personal and sensitive information was anonymized and handled with the highest standards of confidentiality. Compensation was provided uniformly according to the study protocol, ensuring fair treatment throughout the 70-day experience sampling process.

\section*{Data Availability Statement}
The dataset of Flemish daily narratives used in this study contains sensitive personal information and was collected from participants in Europe under the General Data Protection Regulation (GDPR). To protect participant privacy while supporting scientific reproducibility, the anonymized dataset may be made available to qualified researchers for non-commercial research purposes, subject to a data-use agreement. Requests should be directed to the corresponding author (K.H.).

\bibliography{custom}

\newpage
\section*{Appendix}
\appendix

\section{Participants}

Participants were a community sample recruited through flyers, online posts, and word-of-mouth. To be eligible, participants needed to be native Dutch speakers at least 18 years old, living in Belgium, with a smartphone in good condition. Interested individuals were directed to an eligibility survey, and these criteria were later verified during the online introduction session.

Participants were compensated up to \euro250  for completing a 70-day (10-week) experience sampling protocol, including biweekly online surveys. They earned \euro0.50 for each completed experience sampling prompt (4 prompts $\times$ 70 days = 280 prompts or \euro140 max), \euro10 for each short online survey (2, 4, 6, and 8 weeks; \euro40 max), and \euro15 for each long online survey (0 and 10 weeks, \euro30 max). Participants who remained in the study for at least 60 days received a bonus of \euro40. Compensation was provided as a lump sum at the study’s end. To remain in the study, participants needed to complete at least 75\% of prompts and submit verbal descriptions of at least 25 words. Compliance was monitored via periodic checks and summary reports.

A total of 115 participants enrolled (age: 18--65, M = 27.26, SD = 9.86; gender: 58 women, 56 men, 1 other). Of these, 10 chose to leave the study after partial completion, and 3 were dismissed for poor compliance (i.e., response rate under
50\%). The remaining 102 participants ranged in age from 18 to 65 (M = 26.47, SD = 8.87): roughly 14\% of participants were under 20 years old, 63\% between 20 and 30, 15\% 30-40, 5\% 40-50, 3\% 50-60, and 1\% over 60. The sample comprised 52 women (51\%), 49 men (48\%), and 1 other (1\%). Ninety-four (92\%) participants described themselves as having a Flemish, Belgian,
European, or White background, 3 (3\%) described a mixed Belgian and other origin background, 1 (1\%) reported another ethnic background (Chinese), and 4 (4\%) chose not to say.


Study procedures and materials were reviewed and approved by the KU Leuven Social and Societal Ethics Committee (SMEC), protocol G-2023-6379-R3(AMD). Data collection occurred from August 2023 through July 2024, with all study instructions provided in Dutch.

\subsection*{Procedure and Materials}

Participants completed 70 days of experience sampling, receiving four prompts daily via a dedicated smartphone app (\textit{m-Path}) \cite{mestdagh2023mpath}. Prompts were sent pseudo-randomly between 9 AM and 9 PM, with at least an hour between them. At each prompt, participants responded to: “What is going on now or since the last prompt, and how do you feel about it?” Responses were recorded as one-minute voice messages or 3--4 sentence typed responses.

While completing prompts, the \textit{m-Path} app used phone sensors to record GPS coordinates, ambient noise levels, step count, and recent phone app usage (Android devices only, app names recorded). Participants also completed biweekly online surveys assessing well-being and emotional functioning, but these data were not used in the current analysis.

Model evaluations were performed using an NVIDIA RTX-5000 GPU and implemented in PyTorch.

\section{An Overview of the Three Topic Modeling Approaches: KMeans, LDA, and BERTopic}

In this section, we compare and detail the key components of the specific topic modeling approaches chosen for this study: a TF-IDF based KMeans clustering baseline, the probabilistic Latent Dirichlet Allocation (LDA), and the embedding-based BERTopic.

\subsection{KMeans}
To represent a non-probabilistic, geometric clustering baseline, we employ KMeans \cite{macqueen1967some, lloyd1982least, berkhin2006survey, sinaga2020unsupervised}. For vectorization, we use TF-IDF, as it is a standard feature extraction method for classical document clustering and provides a stronger baseline than raw term counts due to its ability to down-weight frequent, uninformative words \cite{salton1988term, ramos2003using}. Following prior work demonstrating its efficacy on Dutch texts \cite{kamiloglu2025what}, we apply KMeans to partition the resulting TF-IDF document vectors, where each cluster centroid represents a latent topic.

We deliberately avoided using embedding-based vectorizers for this baseline to ensure a clear comparison between distinct modeling paradigms. The objective of this study is to contrast an end-to-end, embedding-native model (BERTopic) against established, non-embedding methods. Using embeddings to generate features for KMeans would position it as a hybrid model, which, while a valid approach, falls outside the scope of this direct comparison. 

\subsection{Latent Dirichlet Allocation (LDA)}
As a canonical probabilistic baseline, we include LDA, a traditional probabilistic topic modeling technique that assumes a hierarchical Bayesian structure, where documents are generated by sampling topics from a Dirichlet-distributed prior, with words subsequently sampled from topic-specific multinomial distributions \cite{blei2003latent, shin2025comparison}. It employs CountVectorizer \cite{scikit-learn} to transform preprocessed text into a document-term matrix (DTM), where each row represents a document, and each column corresponds to a unique term.

\subsection{BERTopic}
BERTopic is a relatively recent topic modeling approach 
that leverages transformer-based embeddings to capture nuanced semantic relationships in textual data and generate topics \cite{grootendorst2022bertopic}. The modular, multi-stage pipeline primarily involves three steps: dimensionality reduction, document clustering, and topic extraction.

\begin{figure*}[!htbp]
    \centering
    \includegraphics[width=\linewidth]{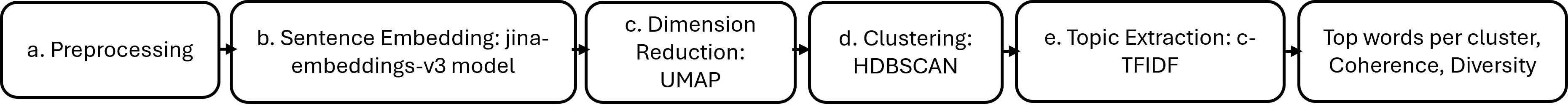}
    \caption{BERTopic implementation pipeline}
    \label{fig:BERTopicpipeline}
\end{figure*}

\subsubsection{Dimensionality Reduction}
To address the challenges posed by the high dimensionality of the embeddings, we employ Uniform Manifold Approximation and Projection (UMAP) \cite{mcinnes2018umap} to reduce the high-dimensional embeddings to a low-dimensional space. UMAP is a non-linear technique adept at preserving both the local and global structure of the data from the high-dimensional space, which is crucial for effective subsequent clustering \cite{grootendorst2022bertopic}.

\subsubsection{Clustering}
The reduced embeddings are then clustered using Hierarchical Density-Based Spatial Clustering of Applications with Noise (HDBSCAN) \cite{campello2013density}. We selected HDBSCAN as the default clustering component of BERTopic for several reasons that make it particularly well-suited for our corpus of open-ended personal narratives.

First, unlike centroid-based algorithms such as KMeans, HDBSCAN does not require the number of topics to be specified \textit{a priori}. This is a significant advantage when exploring a novel dataset where the thematic structure is unknown. Second, HDBSCAN is a density-based algorithm that identifies clusters of varying shapes and densities, which is a better fit for the fluid and unstructured nature of narrative text than the globular cluster assumption of KMeans.  

More crucially, HDBSCAN can classify data points as noise (outliers) rather than forcing every document into a topic. This is invaluable for our dataset, as personal narratives can often be non-topical or too ambiguous to be assigned to a coherent theme. By treating these as noise, we ensure that the resulting topic representations are cleaner and more semantically cohesive. While BERTopic's modularity allows for other clustering algorithms, our use of HDBSCAN is a deliberate choice to align the modeling approach with the inherent characteristics of our data \cite{grootendorst2022bertopic}.

\subsubsection{Topic Extraction}
To provide interpretable representations of each topic, the key terms are extracted using BM25-weighted class-based TF-IDF (c-TF-IDF) \cite{grootendorst2022bertopic}. This method treats all documents within a given cluster as a single composite document. It then calculates term frequencies within this composite document and weighs them by the inverse frequency of the term across all other clusters. This process extracts words that are most representative of a specific topic cluster.

\begin{figure*}[t]
    \centering
    \includegraphics[width=\textwidth]{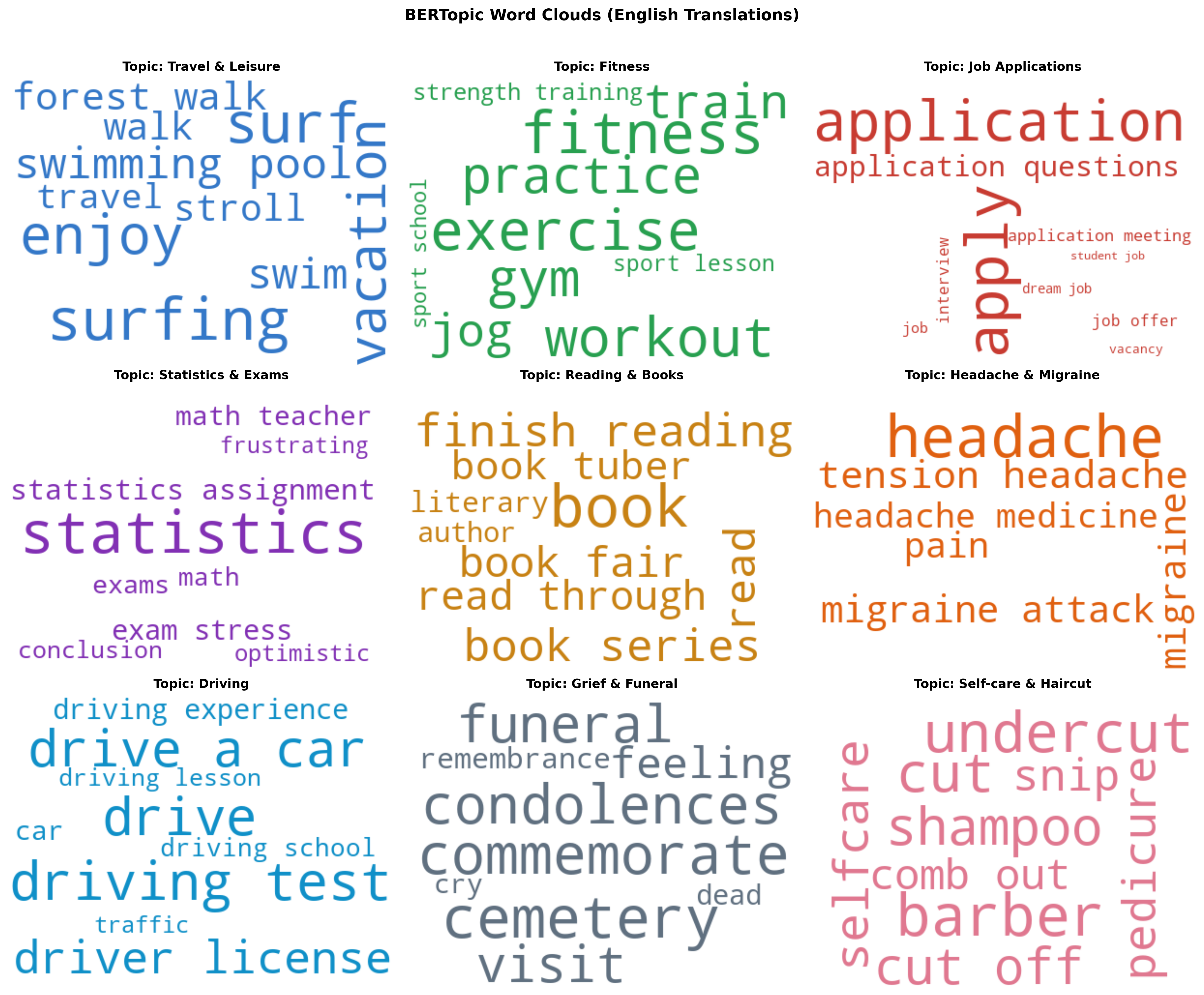}
    \caption{Word clouds of representative BERTopic topics (English translations). 
    Each cloud displays the top words for a selected topic, with font size 
    proportional to word relevance within the topic.}
    \label{fig:wordclouds}
\end{figure*}

\section{Silhouette Analysis}
\begin{figure}[H]
    \centering
    \includegraphics[width=\columnwidth]{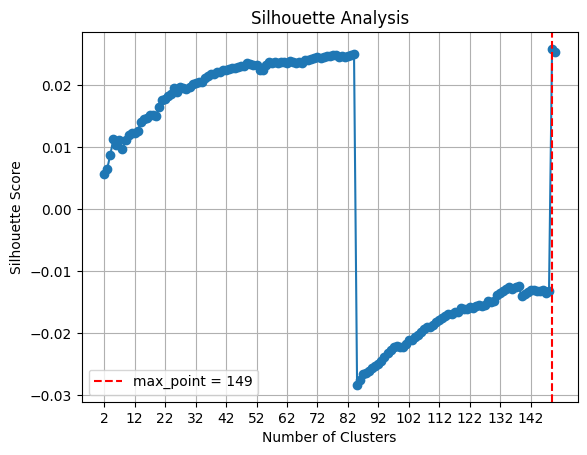}
    \caption{Silhouette Score vs Number of Clusters}
    \label{fig:Silhouette}
\end{figure}

\section{Human Annotation}

\begin{figure*}[!htbp]
    \centering
    \includegraphics[width=\textwidth]{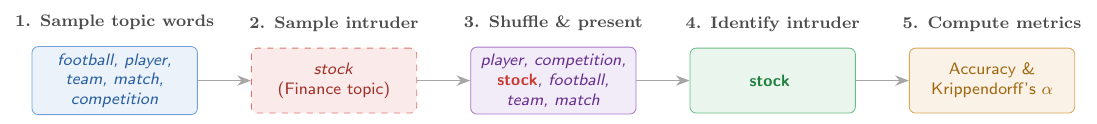}
    \caption{Schematic of the word intrusion task. The intruder word (\textcolor{intruderred}{\textbf{stock}}, shown in red) is sampled from a high-probability term of an unrelated topic (Finance) and inserted at a random position in the six-word set.}
    \label{fig:word-intrusion}
\end{figure*}

\begin{figure}[H]
    \centering
    \includegraphics[width=\columnwidth]{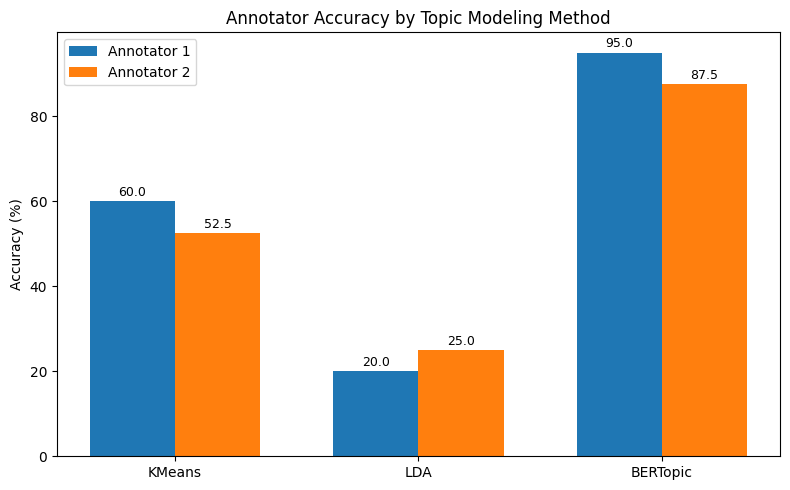}
    \caption{Annotator's Accuracy}
    \label{fig:Accuracy}
\end{figure}

\begin{figure}[H]
    \centering
    \includegraphics[width=\columnwidth]{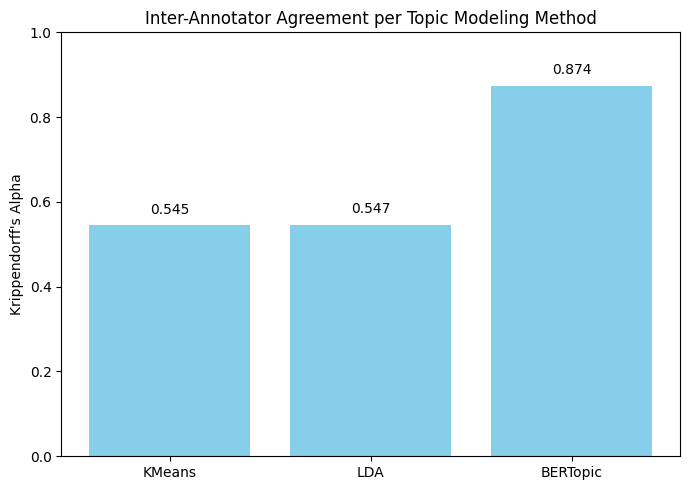}
    \caption{Inter-Annotator Agreement}
    \label{fig:IAA}
\end{figure}

\section{Topic Distribution (BERTopic)
}
We extracted the continuous probability [0,1] that each text loaded onto each topic. To
account for the fact that texts often included content corresponding to multiple themes or
events, we considered a topic to be mentioned if its loading probability surpassed a threshold of
0.01 (i.e., binarizing the data), with this threshold established based on an examination of the
distribution of loading probabilities.
We observed that the 75 topics (excluding -1) were used across a range of texts (min =
24 [0.1\%], max = 20,186 [82\%], M = 1,085 [4\%], SD = 2,311 [9\%]). Likewise, texts typically
touched on multiple topics (min = 0, max = 21 [28\%], M = 4 [6\%], SD = 3 [4\%]).

\section{Word-Intrusion Task Accuracy}
\label{app:ci}

To assess the reliability of the word-intrusion results given the two-annotator
pool, we report 95\% Wilson confidence intervals for each model's accuracy over the
80 judgments per model (40 topics $\times$ 2 annotators). Chance accuracy on the
six-word task is $1/6 \approx 0.167$.

\begin{table}[t]
\centering
\small
\begin{tabular}{lccc}
\toprule
\textbf{Model} & \textbf{Correct} & \textbf{Accuracy} & \textbf{95\% CI} \\
\midrule
BERTopic & 73/80 & 0.912 & [0.830, 0.957] \\
KMeans   & 45/80 & 0.562 & [0.453, 0.666] \\
LDA      & 18/80 & 0.225 & [0.147, 0.328] \\
\bottomrule
\end{tabular}
\caption{Word-intrusion accuracy per model.}
\label{tab:ci-acc}
\end{table}

The three intervals are non-overlapping, confirming the ranking
BERTopic $>$ KMeans $>$ LDA despite the two-annotator pool. LDA's interval lies
just above the chance level, indicating that its topics are close to
indistinguishable from incoherent word sets for human raters, despite LDA
attaining the highest $C_v$ score among the three methods.

\end{document}